\documentclass[11pt]{article}

\usepackage[preprint]{acl}
\usepackage{booktabs}
\usepackage{tabularx}
\usepackage{pgfplots}
\usepackage{colortbl}
\usepackage{adjustbox}
\pgfplotsset{compat=1.17}
\usepackage{xcolor}
\definecolor{k1}{RGB}{246,192,198}
\definecolor{k2}{RGB}{252,228,168}
\definecolor{k3}{RGB}{174,222,186}
\definecolor{k4}{RGB}{168,200,240}
\definecolor{k5}{RGB}{208,184,240}
\definecolor{k6}{RGB}{186,168,224}
\definecolor{panelrule}{RGB}{140,180,220}
\definecolor{panelhead}{RGB}{210,228,246}
\newcommand{\ks}[2]{%
  \mbox{\setlength{\fboxsep}{1.15pt}\colorbox{k#1}{\strut\kern0.35pt#2\kern0.35pt}}%
  \hspace{0.45pt}\allowbreak
}

\usepackage{times}
\usepackage{latexsym}
 \usepackage{amsmath,amssymb}

\usepackage[T1]{fontenc}

\usepackage[utf8]{inputenc}

\usepackage{microtype}

\usepackage{inconsolata}

\usepackage{graphicx}

\title{Next Token or Vector? Learning How Far to Decode}

\newcommand{\aclAuthor}[2]{%
  {\normalfont\fontsize{12pt}{14pt}\selectfont\bfseries
  \mbox{#1\textsuperscript{#2}}}%
}

\author{%
\aclAuthor{Sai Sreenivas Chintha}{1*\textdagger}\hspace{0.4em}%
\aclAuthor{Chidera Biringa}{1*\textdagger}\hspace{0.4em}%
\aclAuthor{Rishab Sharma}{2*}\hspace{0.4em}%
\aclAuthor{Anusha Agrawal}{2*}%
\\[5pt]
\aclAuthor{Prachi Badarayani}{1}\hspace{1em}%
\aclAuthor{Gemma Zhang}{1}\hspace{1em}%
\aclAuthor{Nigel Boachie Kumankumah}{1}%
\\[5pt]
\aclAuthor{Marco Ayala}{1}\hspace{2em}%
\aclAuthor{Sulaiman Vesal}{1}%
\\[7pt]
{\normalfont\fontsize{12pt}{14pt}\selectfont
\textsuperscript{1}Microsoft
\qquad
\textsuperscript{2}University of Massachusetts Amherst
}%
}

\begin{document}
\maketitle

\begingroup
\renewcommand{\thefootnote}{\fnsymbol{footnote}}

\footnotetext[1]{%
  \fontsize{6.5pt}{8pt}\selectfont
  Equal contribution
}

\footnotetext[2]{%
  \fontsize{6.5pt}{8pt}\selectfont
  \mbox{Corresponding authors:
  \texttt{\{schintha, cbiringa\}@microsoft.com}}%
}

\endgroup

\begin{abstract}

Standard autoregressive decoding advances one token at a time, even when several upcoming tokens can be predicted together. We propose CADENCE (\textbf{C}ontext-\textbf{A}daptive \textbf{DE}coding), a generation method that learns both what to draft and how far to draft. Gaussian prediction heads represent future spans as continuous latent vectors, and a shared decoder converts the selected vector into token predictions in parallel. A learned router selects the prediction length from the model’s own hidden states and predicted Gaussian parameters, adapting generation granularity to the current context with single-token generation as a fallback. The backbone verifies each proposed continuation and retains the prefix that matches its greedy predictions. Experiments with Llama-3.1-8B-Instruct on Alpaca, GSM8K, and HumanEval show that router-selected positions yield higher draft acceptance than applying the corresponding prediction length everywhere. CADENCE achieves up to 1.3× higher generation throughput than standard greedy decoding.

\end{abstract}

\section{Introduction}


Large language models generate text autoregressively, predicting one token at a time from the preceding context. This sequential process makes decoding latency grow with response length, increasing user wait time and limiting the number of requests a serving system can process with fixed computational resources. Faster inference can therefore improve both responsiveness and serving capacity. However, accelerating generation remains difficult because every accepted token must remain consistent with the preceding context. Token predictability also varies substantially across a sequence. Context strongly constrains common phrases and copied spans, while other continuations require recall or computation. For example, in \textit{``What is the square root of 7? The square root of 7 is 2.646,''} the repeated context makes the second occurrence of ``7'' easy to predict, whereas producing ``2.646'' requires recalling or computing the value~\citep{leviathan2024looking}. Standard decoding applies the same backbone computation in both cases. Adaptive decoding can instead take short steps when uncertainty is high and longer steps when the context supports several upcoming tokens.

\begin{figure}[t]
\centering
{\footnotesize
\ks{1}{$a{=}1$}%
\ks{2}{$a{=}2$}%
\ks{3}{$a{=}3$}%
}

\vspace{4pt}

\fcolorbox{panelrule}{white}{%
\begin{minipage}{\dimexpr\linewidth-2\fboxsep-2\fboxrule\relax}
\colorbox{panelhead}{\parbox{\dimexpr\linewidth-2\fboxsep\relax}{%
  \small\bfseries What is the square root of 7?}}
\vspace{2pt}
{\small\raggedright
\ks{2}{The square}%
\ks{3}{ root of }%
\ks{3}{7 is a}%
\ks{1}{ non}%
\ks{1}{-}%
\ks{1}{integer}%
\ks{1}{ value}%
\ks{1}{,}%
\ks{1}{ approximately}%
\ks{1}{ equal}%
\ks{2}{ to }%
\ks{2}{2.}%
\ks{1}{645}%
\ks{1}{751}%
\ks{1}{311}%
\ks{1}{.}%
\par}
\end{minipage}}

\caption{Variable-length generation for ``What is the square root of 7?''. Here, 'a' is the acceptance length under verification. The opening and ``root of 7 is a'' are multi-token accepts; the digits of the approximation fall back to one token at a time.}
\label{fig:sqrt7}
\end{figure}

Existing methods accelerate generation by drafting multiple tokens for parallel verification or by changing the representation used during generation. They construct drafts from auxiliary token heads, internal model features, diffusion blocks, or compressed latent representations~\citep{leviathan2023speculative,cai2024medusa,li2024eagle,li2025eagle3,pmlr-v306-chen26bv,shao2025calm,tan2025colar}. These approaches show that models can advance by more than one token per decoding step, but the chosen representation also determines what kinds of length-control decisions the model can make.


Adaptive methods already control draft length through learned stopping and dynamic tree construction~\citep{huang2024specdec++,liu2026talon,wang2026speculative}. Recent serving work further shows that indiscriminately verifying long drafts can waste target-model batch capacity when acceptance decays, particularly under high concurrency~\cite{cheng2026dspark}. Adaptive verification addresses this cost by pruning low-value draft suffixes before target verification. We ask whether the model can move this decision earlier by selecting among latent representations of different candidate span lengths before decoding the proposed span.

We propose \textbf{CADENCE}, which learns how far to draft from horizon-specific Gaussian representations of candidate spans. At each generation step, Llama-3.1-8B-Instruct produces a context representation and predicts the immediate next token. From this context, Gaussian heads predict one fixed-dimensional latent for each candidate horizon. The router combines the context with the predicted Gaussian parameters to select a horizon or fall back to the backbone token. The shared Perceiver-style decoder then maps the mean of the selected Gaussian to parallel token predictions~\citep{jaegle2022perceiver}. The backbone verifies the resulting proposal and continues from the accepted prefix.

We test whether learned routing selects contexts in which longer span predictions succeed. Using the same span predictors and decoder, we compare router-selected generation with fixed-horizon generation. We then measure whether the Gaussian parameters improve routing beyond the backbone context alone. Finally, we compare the complete system's generation throughput with standard greedy backbone decoding.

\begin{figure*}[h]
    \centering
    \includegraphics[width=0.9\linewidth]{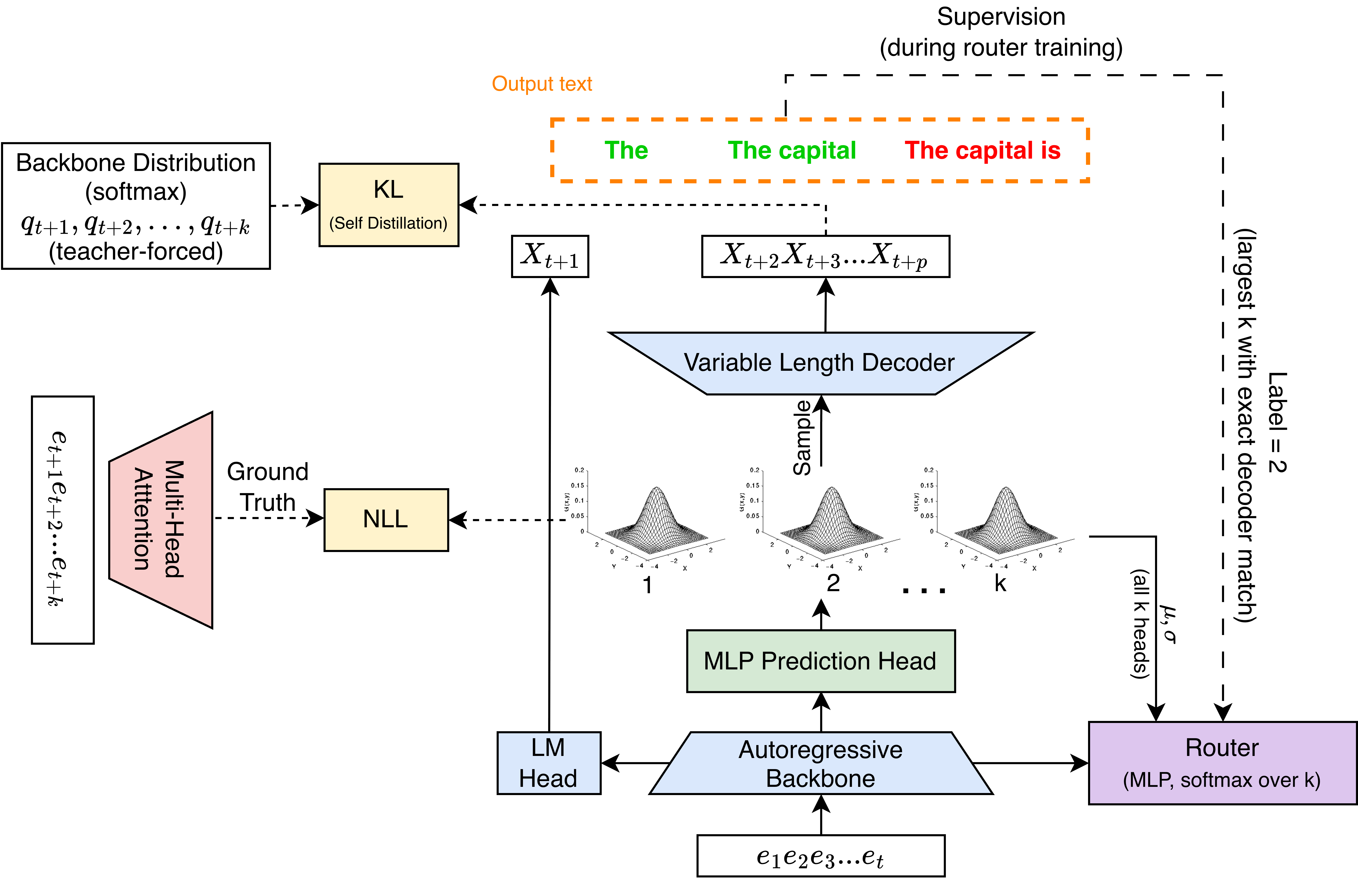}
\caption{Overview of the CADENCE architecture and training framework. The backbone provides context representations for Gaussian heads that predict future-span latents. A shared decoder generates token predictions, while a router selects the prediction horizon using the context and Gaussian parameters. Training uses Gaussian negative log-likelihood (NLL), KL self-distillation, and full-span matching supervision for the router.}
    \label{fig:architecture}
\end{figure*}

\begin{table}[t]
\centering
\small
\setlength{\tabcolsep}{3pt}
\renewcommand{\arraystretch}{1.15}
\caption{Prediction forms and length-control mechanisms.}
\label{tab:related_work}
\begin{tabularx}{\columnwidth}{
    @{}ll>{\raggedright\arraybackslash}X@{}
}
\toprule
\textbf{Method} & \textbf{Prediction} & \textbf{Length control} \\
\midrule
CALM      & Chunk vector     & Fixed chunk size \\
CoLaR     & Reasoning latent & Prompted compression \\
MEDUSA    & Token tree       & Configured tree \\
EAGLE-3   & Token tree       & Confidence-based tree selection \\
DFlash    & Token block      & Configured block size \\
SpecDec++ & Token chain      & Learned stopping \\
AdaEDL    & Token chain      & Entropy-based stopping \\
TALON     & Token tree       & Confidence-gated expansion \\
Re-SpS    & Token tree       & Learned tree configuration \\
\textbf{CADENCE}
          & \textbf{Span latent}
          & \textbf{Learned span selection} \\
\bottomrule
\end{tabularx}
\end{table}

\section{Related Work}
We organize prior work by generation process, predicted representation,
and draft-length control. Table~\ref{tab:related_work} summarizes the
key distinctions among the closest methods.

\subsection{Diffusion and Insertion Models}
SEDD, MDLM, and LLaDA generate text through reverse discrete diffusion~\citep{pmlr-v235-lou24a,sahoo2024simple,nie2026large}. Block Diffusion combines autoregression across blocks with diffusion within each block~\citep{arriola2025block}. The Insertion Transformer predicts tokens at multiple insertion locations~\citep{pmlr-v97-stern19a}. CADENCE instead retains causal backbone decoding and proposes variable-length continuations from the accepted token prefix.

\subsection{Continuous Representations}
CALM compresses fixed-size token chunks into vectors and predicts subsequent vectors autoregressively~\citep{shao2025calm}. CoLaR predicts Gaussian distributions over compressed reasoning embeddings and controls compression through a prompt-specified factor~\citep{tan2025colar}. Both methods feed their latent representations into subsequent backbone steps. CADENCE instead decodes the selected span latent into parallel token predictions and continues generation from the prefix accepted by the backbone.

\subsection{Speculative and Adaptive Decoding}
Speculative decoding combines inexpensive drafting with target-model verification~\citep{leviathan2023speculative}. Parallel token heads can produce these drafts directly~\citep{gloeckle2024better,cai2024medusa}. EAGLE and EAGLE-3 instead draft autoregressively from target-model features~\citep{li2024eagle,li2025eagle3}, while DFlash predicts a masked token block in parallel using injected target features~\citep{pmlr-v306-chen26bv}.

SpecDec++ and AdaEDL stop sequential drafting using learned acceptance estimates and entropy-based criteria, respectively~\citep{huang2024specdec++,pmlr-v262-agrawal24a}. EAGLE-3 and TALON use confidence to adapt draft trees~\citep{li2025eagle3,liu2026talon}, while Re-SpS learns tree configurations from target-model hidden states~\citep{wang2026speculative}. DSpark adapts verification length using predicted prefix-survival probabilities and a hardware throughput profile~\citep{cheng2026dspark}. CADENCE instead selects among horizon-specific Gaussian span latents before decoding and conditions this decision on the backbone context and predicted Gaussian parameters. Backbone verification then determines the accepted prefix.

\section{Methodology}
\label{sec:method}

CADENCE augments a pretrained autoregressive language model with horizon-specific Gaussian prediction heads, a shared Perceiver IO-inspired decoder~\citep{jaegle2022perceiver}, and a learned router. At each generation step, the backbone predicts the next token and produces a context representation from which the Gaussian heads predict latent representations for different prediction horizons. Using the context and Gaussian parameters, the router selects a horizon $k \in \{1,\ldots,5\}$ or a backbone-only fallback. For the selected horizon, the shared decoder predicts $k$ additional tokens in parallel, forming a draft of length $k+1$ including the backbone token. Greedy verification against the frozen backbone commits the longest matching prefix, with accepted length $1 \leq a \leq k+1$. The fallback commits only the backbone token. Figure~\ref{fig:architecture} illustrates the architecture.


\subsection{Multi-step Gaussian heads}
\label{sec:gaussians}

We make a simple modeling choice by representing the $k$ future tokens as a continuous diagonal Gaussian distribution.

Let $h_t^{(\ell)}$ denote the hidden state at position $t$ in backbone layer $\ell$. We construct the context representation by concatenating the last-token states from the final three layers and projecting the result back to the backbone width:
\begin{equation}
c_t
=
\mathrm{Proj}\!\left(
\left[h_t^{(-3)};\,h_t^{(-2)};\,h_t^{(-1)}\right]
\right)
\end{equation}
Using several final-layer states exposes the prediction heads to complementary representations from the backbone's latest processing stages. The linear projection gives every downstream component a common input width and prevents the concatenation from increasing their parameter cost.

A shared MLP trunk processes $c_t$, and one horizon-specific head predicts the parameters of a $D$-dimensional diagonal Gaussian,
\begin{equation}
q_{t,k}(z)
=
\mathcal{N}\!\left(
z;\mu_{t,k},
\operatorname{diag}(\sigma_{t,k}^{2})
\right)
\end{equation}
The shared trunk captures information useful across span lengths, while the separate heads allow each horizon to learn its own latent distribution. Diagonal covariance keeps the covariance parameterization linear in $D$ and makes likelihood training tractable at every generation position. We use $D=512$ for all horizons, holding latent capacity constant as span length changes.


The regression target compresses the embeddings of the $k$ future tokens into one span representation. Mean pooling discards token order, while directly adding positional vectors before averaging entangles position with the semantic values being compressed. We instead use a learned multi-head attention pooler and apply rotary position embeddings to its queries and keys~\citep{su2024roformer}. This design lets attention depend on token position while leaving the value vectors unchanged. The pooler then linear down-projects the attended representation:
\begin{equation}
z^{\star}_{t,k}
= \mathrm{DownProj}\!\left(\mathrm{MHA}\!\left(e_{t+2:t+1+k}\right)\right)
\end{equation}
where $e_i$ is the backbone embedding of the ground-truth token at position $i$. The span begins at $t+2$ because the backbone language-model head predicts position $t+1$ directly; Section~\ref{sec:self-distill} formalizes this division. We pretrain the pooler with the decoder and then freeze it, making $z^{\star}_{t,k}$ a fixed supervision target during Gaussian-head training.

We train each Gaussian head by minimizing the negative log-likelihood of its pooled span target:

\begin{equation}
\small
\mathcal{L}_{\mathrm{NLL},k}
= \frac{1}{2}\sum_{d=1}^{D}
\left[
\log\!\left(2\pi\sigma_{t,k,d}^{2}\right)
+ \frac{(z^{\star}_{t,k,d}-\mu_{t,k,d})^{2}}
{\sigma_{t,k,d}^{2}}
\right]
\end{equation}

\subsection{Perceiver IO Decoder}
\label{sec:decoder}


The decoder follows the cross-attention design of Perceiver IO~\citep{jaegle2022perceiver}. For horizon $k$, it uses the selected latent $z\in\mathbb{R}^{D}$ as a shared key and value and assigns one learned positional query to each of the $k$ output positions. Four cross-attention layers let these queries extract position-specific information from the shared span latent. A linear output map then produces all $k$ vocabulary distributions in parallel:

\begin{equation}
p_{\theta}(\,\cdot \mid z, k) = \mathrm{softmax}\!\left(\mathrm{Decoder}(z, k)\right) \in \mathbb{R}^{k \times V}
\end{equation}


All horizons share the decoder parameters and positional query bank. The selected horizon determines how many queries participate in decoding. Sharing the decoder forces the Gaussian heads to produce compatible latent representations and avoids training a separate token decoder for every span length.

\subsection{Router}
\label{sec:router}

The Gaussian heads represent candidate spans, but the model must still decide how far to draft from the current context. The router makes this decision with a residual MLP of width 512. It conditions each decision on the backbone context $c_t$ and the Gaussian parameters $(\mu_{t,k},\sigma_{t,k})$ predicted for all candidate horizons. These inputs expose both the contextual evidence for continuing and the latent distributions associated with each possible span length. The router outputs a softmax over six classes: five prediction horizons and a backbone-only fallback.

\subsection{Training}
\label{sec:training}


We train CADENCE in three stages that isolate the three functions required for adaptive span prediction. First, we pretrain the attention pooler and shared decoder to establish a latent space from which the decoder can reconstruct future spans. Second, with the backbone and pooler frozen, we jointly train the Gaussian heads, context projection, and decoder to predict those span representations from context and recover the backbone's future-token distributions. Third, we freeze the span-prediction modules and train the router to select the longest appropriate horizon. This schedule prevents the router from learning against continuously changing draft behavior and separates representation learning, contextual span prediction, and horizon selection.



\subsubsection{Decoder and pooler pretraining}
\label{sec:decoder-pretrain}

Decoder pretraining does not involve the Gaussian heads or the distillation loss. The pooler compresses the ground-truth embeddings of each span, and the decoder reconstructs the span with token-level cross-entropy,
\begin{equation}
\mathcal{L}_{\mathrm{dec}}
= -\sum_{k}\sum_{m=0}^{k-1}
\log p_{\theta}\!\left(y_{t+2+m} \mid z^{\star}_{t,k}, k\right)_{m}
\end{equation}
Training the decoder directly from the pooled target first establishes that the latent space supports reconstruction before the Gaussian heads must predict points in that space from context. This separation prevents decoder failures from being conflated with Gaussian prediction errors. We train both the pooler and decoder during this stage and then freeze the pooler, fixing the target $z^{\star}_{t,k}$ used by the Gaussian negative log-likelihood.



\subsubsection{Joint training of the Gaussian heads}
\label{sec:joint}
\label{sec:self-distill}


After establishing the span latent space, we freeze the backbone and pooler and jointly train the context projection, Gaussian heads, and decoder. The objective combines Gaussian negative log-likelihood, which aligns each predicted latent distribution with its pooled span target, and token-level self-distillation, which trains the decoder to reproduce the backbone's predictive distribution.


The self-distillation term aligns training with verification. Greedy verification evaluates the draft against the backbone, so the decoder learns from the same backbone distributions that later determine whether its tokens are accepted. This follows the broader use of frozen target-model supervision for auxiliary draft predictors~\citep{cai2024medusa,li2024eagle}. Let $q_t=\mathrm{softmax}(\mathrm{LM}(h_t))$ denote the backbone distribution at position $t$. The decoder receives a reparameterized sample
$z_{t,k}=\mu_{t,k}+\sigma_{t,k}\odot\epsilon$,
where $\epsilon\sim\mathcal{N}(0,I)$, and averages the divergence across the $k$ decoded positions:
\begin{equation}
\mathcal{L}_{\mathrm{KL},k}
= \frac{1}{k}\sum_{m=0}^{k-1}
\mathrm{KL}\!\left(
q_{t+1+m}
\;\middle\|\;
p_{\theta}(\,\cdot \mid z_{t,k}, k)_{m}
\right)
\end{equation}
The second term is the negative log-likelihood $\mathcal{L}_{\mathrm{NLL},k}$ of the pooled target under the predicted Gaussian. The joint objective is
\begin{equation}
\mathcal{L}
= \lambda_{\mathrm{NLL}}\sum_{k}\mathcal{L}_{\mathrm{NLL},k}
+ \lambda_{\mathrm{KL}}\sum_{k}\mathcal{L}_{\mathrm{KL},k}
\end{equation}
We set $\lambda_{\mathrm{KL}}=1$ and $\lambda_{\mathrm{NLL}}=0.005$. The distillation term receives unit weight because it directly supervises the token distributions evaluated during verification. Each Gaussian NLL sums over $D=512$ latent dimensions, and the objective then aggregates this loss across five horizons. We therefore downweight the NLL term to balance its scale against token-level distillation while retaining direct supervision of both $\mu_{t,k}$ and $\sigma_{t,k}$.

\subsubsection{Router training}
\label{sec:router-train}




We train the router after freezing the backbone, pooler, Gaussian heads, and decoder, ensuring that every label refers to fixed draft behavior. For each supervised position, we decode every horizon and mark horizon $k$ as correct when all $k$ predicted tokens match the corresponding backbone targets. We assign the largest correct horizon as the training label because it yields the greatest valid progress from that context. If no horizon matches completely, we assign class $0$, which represents backbone-only generation.

Longer horizons produce fewer full-span matches, creating an imbalanced six-class problem. We use multiclass focal loss~\citep{lin2018focallossdenseobject} with focusing parameter $\gamma=1$:
\begin{equation}
\mathcal{L}_{\mathrm{focal}}
= -\alpha_{y}\,(1 - p_{y})\log p_{y}
\end{equation}
where $p_{y}$ is the softmax probability of the labeled class. We set each class weight in proportion to the inverse square root of its training frequency, $\alpha_c\propto n_c^{-1/2}$. This weighting increases the contribution of rare long-horizon labels while remaining less aggressive than inverse-frequency weighting.


\subsection{Inference}
\label{sec:inference}

At each generation step, CADENCE computes the context representation $c_t$ and the Gaussian parameters for all candidate horizons. The router then selects a horizon $k\in\{1,\ldots,5\}$ or the backbone-only fallback. For a selected horizon, the decoder uses the Gaussian mean $\mu_{t,k}$ as its latent. Using the mean makes routing and decoding deterministic and matches the representations used to construct router labels. The proposed draft is
\begin{align*}
    &\big(\hat{y}_{t+1},\, \hat{y}_{t+2:t+1+k}\big) \\
    &=
    \big(
    \arg\max q_t,\;
    \arg\max p_{\theta}(\,\cdot \mid \mu_{t,k}, k)
    \big)
\end{align*}
where $q_t$ is the backbone's next-token distribution. The proposal contains the backbone token followed by $k$ decoder tokens, giving a total length of $k+1$. If the router selects the fallback class, CADENCE commits only the backbone token.

The proposed draft undergoes greedy verification against the backbone before the accepted tokens are committed to the output sequence.

\section{Experimental Setup}

\subsection{Implementation Details}
\label{sec:implementation_details}

\paragraph{Model Configuration.}
We use Llama-3.1-8B-Instruct as the frozen backbone for all experiments~\citep{grattafiori2024llama3}. We construct the context representation by concatenating the last-token hidden states from the final three backbone layers and projecting them to the backbone hidden dimension. CADENCE uses five Gaussian prediction heads for horizons $k\in\{1,\ldots,5\}$, each producing a 512-dimensional latent representation. A shared Perceiver-style decoder maps the selected latent to future-token predictions, while a residual MLP router with hidden dimension 512 selects the generation horizon from the context representation and predicted Gaussian parameters $(\mu,\sigma)$.

\paragraph{Training Configuration.}
Training proceeds in three stages. First, the attention-based pooler and decoder are pretrained using AdamW with a learning rate of $3 \times 10^{-4}$, weight decay of $0.01$, and cosine learning-rate decay. Next, we jointly train the Gaussian prediction heads and decoder for two epochs with a learning rate of $10^{-4}$ and 1,000 warmup steps. We weight the self-distillation KL divergence and Gaussian negative log-likelihood by $1$ and $0.005$, respectively. Finally, the router is trained for one epoch with a learning rate of $3 \times 10^{-4}$ and 300 warmup steps, using focal loss ($\gamma=1$) and inverse-square-root class-frequency weights. The joint Gaussian-decoder and router stages use an effective batch size of 16 and clip the gradient norm at $1.0$.

\subsection{Datasets}
\label{sec:datasets}

\paragraph{Training Data.}
We use an UltraChat-ShareGPT dataset generated with Llama-3.1-8B-Instruct~\citep{zhuyksir2026ultrachatsharegpt} to train the Gaussian prediction heads, shared decoder, and step-size router.

\paragraph{Evaluation Benchmarks.}
We evaluate CADENCE across three benchmarks covering distinct generation tasks: Alpaca~\citep{alpaca} for instruction-following and free-form text generation, GSM8K~\citep{cobbe2021gsm8k} for mathematical reasoning, and HumanEval~\citep{chen2021evaluating} for code generation. These benchmarks allow us to examine adaptive generation behavior across natural language, structured reasoning, and programming contexts.

\subsection{Evaluation Metrics}
\label{sec:evaluation_metrics}

We evaluate CADENCE using metrics that characterize multi-token generation, draft acceptance, and inference efficiency.

\noindent \textbf{Token Acceptance Rate.}
We compute token acceptance rate as the total number of tokens retained after verification divided by the total number of tokens proposed in the corresponding drafts.

\noindent \textbf{Full-Draft Acceptance Rate.}
We compute full-draft acceptance rate as fraction of proposed drafts for which verification accepts all $k+1$ tokens.

\noindent \textbf{Mean Accepted Length $\tau$.}
We compute the average number of tokens committed per verified drafting step. Unlike token acceptance rate, this metric measures the actual generation progress made at each step.

\noindent \textbf{Speedup.}
We measure generation throughput in tokens per second and report speedup relative to standard autoregressive decoding with the same Llama-3.1-8B-Instruct backbone.

\begin{figure}[t]
  \centering
  \includegraphics[width=\linewidth]{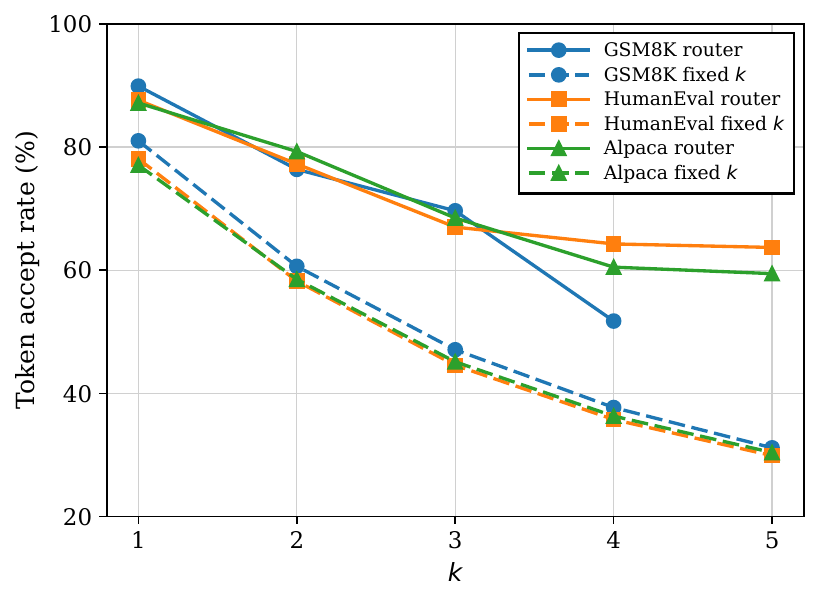}
\caption{Token acceptance rates by prediction horizon under greedy verification on GSM8K, HumanEval, and Alpaca. Solid lines report positions where the router selected each horizon; dashed lines report the corresponding horizon applied at every evaluated position.}
  \label{fig:router-vs-fixedk}
\end{figure}

\section{Results}
\label{sec:results}

We investigate whether \textbf{CADENCE} can adapt generation granularity across different linguistic contexts and tasks. We first evaluate the effectiveness of learned step-size selection and characterize the resulting generation behavior across tasks. We then analyze the contribution of Gaussian prediction parameters to routing decisions and evaluate generation quality and inference efficiency. Unless otherwise specified, all experiments use the router conditioned on both the backbone context representation and the predicted Gaussian parameters. 

\begin{figure}[h]
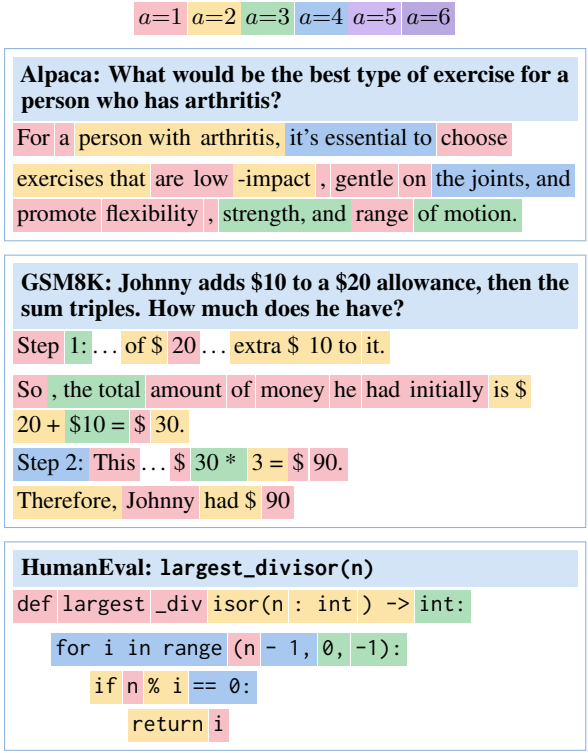

\centering
{\footnotesize
\ks{1}{$a{=}1$}%
\ks{2}{$a{=}2$}%
\ks{3}{$a{=}3$}%
\ks{4}{$a{=}4$}%
\ks{5}{$a{=}5$}%
\ks{6}{$a{=}6$}%
}

\vspace{4pt}

\fcolorbox{panelrule}{white}{%
\begin{minipage}{\dimexpr\linewidth-2\fboxsep-2\fboxrule\relax}
\colorbox{panelhead}{\parbox{\dimexpr\linewidth-2\fboxsep\relax}{%
  \small\bfseries Alpaca: What would be the best type of exercise for a person who has arthritis?}}
\vspace{2pt}
{\small\raggedright
\ks{1}{For}%
\ks{1}{ a}%
\ks{2}{ person with}%
\ks{2}{ arthritis,}%
\ks{4}{ it's essential to}%
\ks{1}{ choose}%
\ks{2}{ exercises that}%
\ks{1}{ are}%
\ks{1}{ low}%
\ks{2}{-impact}%
\ks{1}{,}%
\ks{1}{ gentle}%
\ks{1}{ on}%
\ks{4}{ the joints, and}%
\ks{1}{ promote}%
\ks{1}{ flexibility}%
\ks{1}{,}%
\ks{3}{ strength, and}%
\ks{1}{ range}%
\ks{3}{ of motion.}%
\par}
\end{minipage}}

\vspace{4pt}

\fcolorbox{panelrule}{white}{%
\begin{minipage}{\dimexpr\linewidth-2\fboxsep-2\fboxrule\relax}
\colorbox{panelhead}{\parbox{\dimexpr\linewidth-2\fboxsep\relax}{%
  \small\bfseries GSM8K: Johnny adds \$10 to a \$20 allowance, then the sum triples. How much does he have?}}
\vspace{2pt}
{\small\raggedright
\ks{1}{Step}%
\ks{3}{ 1:}%
\ldots{}%
\ks{2}{ of \$}%
\ks{1}{20}%
\ldots{}%
\ks{2}{ extra \$}%
\ks{2}{10 to}%
\ks{2}{ it.}\\[1pt]
\ks{1}{So}%
\ks{3}{, the total}%
\ks{1}{ amount}%
\ks{1}{ of}%
\ks{1}{ money}%
\ks{1}{ he}%
\ks{1}{ had}%
\ks{1}{ initially}%
\ks{2}{ is \$}%
\ks{2}{20 +}%
\ks{3}{ \$10 =}%
\ks{1}{\$}%
\ks{2}{30.}\\[1pt]
\ks{4}{Step 2:}%
\ks{1}{ This}%
\ldots{}%
\ks{1}{\$}%
\ks{3}{30 * }%
\ks{2}{3 =}%
\ks{1}{ \$}%
\ks{1}{90.}\\[1pt]
\ks{2}{Therefore,}%
\ks{1}{ Johnny}%
\ks{2}{ had \$}%
\ks{1}{90}%
\par}
\end{minipage}}

\vspace{4pt}

\fcolorbox{panelrule}{white}{%
\begin{minipage}{\dimexpr\linewidth-2\fboxsep-2\fboxrule\relax}
\colorbox{panelhead}{\parbox{\dimexpr\linewidth-2\fboxsep\relax}{%
  \small\bfseries HumanEval: \texttt{largest\_divisor(n)}}}
\vspace{2pt}
{\footnotesize\ttfamily
\ks{1}{def}%
\ks{1}{ largest}%
\ks{1}{\_div}%
\ks{2}{isor(n}%
\ks{2}{: int}%
\ks{2}{) ->}%
\ks{3}{ int:}\\[1pt]
\hspace*{1.6em}%
\ks{4}{for i in range}%
\ks{1}{(n}%
\ks{4}{ - 1,}%
\ks{3}{ 0,}%
\ks{3}{ -1):}\\[1pt]
\hspace*{3.2em}%
\ks{2}{if}%
\ks{1}{ n}%
\ks{2}{ \% i}%
\ks{4}{ == 0:}\\[1pt]
\hspace*{4.8em}%
\ks{2}{return}%
\ks{1}{ i}%
\par}
\end{minipage}}

\caption{Examples of variable-length generation across Alpaca, GSM8K, and HumanEval. Colors indicate accepted lengths after greedy verification. Sentence openings and bare amounts are short; templates, prompt copies, and code syntax are longer.}
\label{fig:qual_examples}
\end{figure}

\subsection{Effectiveness of Adaptive Step-Size Selection}
\label{sec:adaptive_selection}

We first examine whether CADENCE identifies contexts suitable for longer generation steps. Figure~\ref{fig:router-vs-fixedk} compares token acceptance rates under adaptive and fixed-horizon generation using greedy verification. For each horizon $k \in \{1,\ldots,5\}$, the adaptive setting evaluates positions where the router selects $k$, whereas fixed-horizon setting applies same $k$ at every generation step. Both settings use the same Gaussian prediction heads and decoder, allowing us to examine the effect of context-dependent horizon selection across domains.

Across all three benchmarks, token acceptance rates generally decrease as the prediction horizon increases. However, router-selected positions consistently exhibit higher acceptance rates than fixed-horizon generation at the corresponding horizon. At $k=4$, for example, CADENCE achieves token acceptance rates of 51.7\%, 64.3\%, and 60.5\% on GSM8K, HumanEval, and Alpaca, respectively, compared with 37.7\%, 35.8\%, and 36.3\% under fixed-horizon generation. This pattern is also observed at shorter horizons, with improvements of approximately 9--10 percentage points at $k=1$ across all three benchmarks. These results indicate that the learned routing policy identifies contexts where the corresponding multi-token predictions are more likely to succeed. Since acceptance is measured conditionally on router-selected positions, the comparison evaluates the effectiveness of horizon selection rather than overall generation efficiency.


\subsection{Generation Granularity Across Tasks}
\label{sec:generation_granularity}

\begin{figure}[h]
    \centering
    \includegraphics[width=\columnwidth]{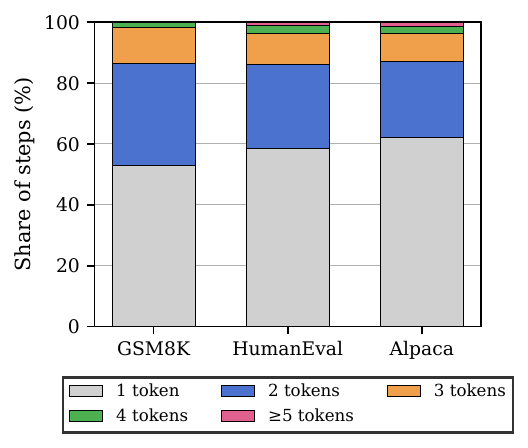}
    \caption{Share of generation steps by the number of tokens retained after greedy verification for the router. One-token steps include both base-model fallbacks and verified drafts for which only one token is retained.}
    \label{fig:accepted_length_share}
\end{figure}

We next examine how CADENCE varies its generation granularity across different tasks. Figure~\ref{fig:accepted_length_share} presents the distribution of accepted lengths on Alpaca, GSM8K, and HumanEval. The distribution includes all generation steps, including backbone-only fallbacks and verified multi-token drafts. While single-token steps remain the most frequent outcome, CADENCE also successfully advances by multiple tokens across all three benchmarks. These results demonstrate that adaptive generation produces variable-length continuations in practice.

Figure~\ref{fig:qual_examples} provides a qualitative view of how generation granularity varies within individual responses. In these examples, longer accepted steps appear in structured or formulaic continuations, such as ``it's essential to'' in Alpaca, ``Step 2:'' in GSM8K, and ``for i in range'' in HumanEval. In contrast, shorter steps appear around sentence beginnings (``For'', ``So''), content-specific words (``choose'', ``gentle''), and numerical answers (``90''). These patterns are consistent with the hypothesis that predictable continuations can be generated at coarser granularity, while less predictable regions may require finer-grained generation.

\begin{table*}[t]
    \centering
    \small
    \setlength{\tabcolsep}{5pt}
    \caption{Effect of Gaussian prediction parameters on adaptive routing on Alpaca under greedy verification. Per-horizon metrics are computed over router-selected drafts; overall metrics include backbone-only fallback steps. Bold indicates the better result.}
    \label{tab:gaussian_router_input}

    \begin{tabular}{llcccccc}
        \toprule
        & & \multicolumn{3}{c}{\textbf{Context-only router}}
            & \multicolumn{3}{c}{\textbf{Gaussian-enhanced router}} \\
        \cmidrule(lr){3-5}
        \cmidrule(lr){6-8}

        \textbf{Horizon}
        & \textbf{Draft}
        & \textbf{$\tau$}
        & \textbf{Token acc.}
        & \textbf{Full acc.}
        & \textbf{$\tau$}
        & \textbf{Token acc.}
        & \textbf{Full acc.} \\
        \midrule

        $k=1$ & 2 & 1.70 & 84.8\% & 69.5\%
              & \textbf{1.74} & \textbf{87.1\%} & \textbf{74.3\%} \\
        $k=2$ & 3 & 2.27 & 75.5\% & 49.8\%
              & \textbf{2.38} & \textbf{79.3\%} & \textbf{58.6\%} \\
        $k=3$ & 4 & 2.68 & 67.1\% & 35.4\%
              & \textbf{2.74} & \textbf{68.5\%} & \textbf{40.0\%} \\
        $k=4$ & 5 & 2.85 & 56.9\% & 21.5\%
              & \textbf{3.02} & \textbf{60.5\%} & \textbf{31.3\%} \\
        $k=5$ & 6 & 3.27 & 54.6\% & 25.1\%
              & \textbf{3.57} & \textbf{59.4\%} & \textbf{30.7\%} \\

        \midrule
        \multicolumn{8}{l}{\textit{Overall generation}} \\
        \addlinespace[2pt]

        \multicolumn{2}{l}{Multi-token drafting steps (\%)}
            & \multicolumn{3}{c}{47.1\%}
            & \multicolumn{3}{c}{\textbf{50.0\%}} \\

        \multicolumn{2}{l}{Mean Accepted Length ($\tau$)}
            & \multicolumn{3}{c}{1.50}
            & \multicolumn{3}{c}{\textbf{1.56}} \\

        \bottomrule
    \end{tabular}
\end{table*}

\subsection{Effectiveness of Gaussian-Enhanced Router}
\label{sec:gaussian_ablation}
We next examine whether the predicted Gaussian parameters provide useful information for prediction horizon selection beyond the backbone context representation. We compare a context-only router with a Gaussian-enhanced router conditioned on both the context representation and predicted Gaussian parameters $(\mu,\sigma)$. Both configurations use the same Gaussian prediction heads and decoder, differing only in information available to the router. Table~\ref{tab:gaussian_router_input} reports their performance on Alpaca.

Incorporating Gaussian parameters improves all three per-horizon metrics across the evaluated horizons. At $k=4$, token acceptance increases from 56.9\% to 60.5\%, while full-draft acceptance improves from 21.5\% to 31.3\%. At $k=5$, mean accepted length increases from 3.27 to 3.57 tokens. At the generation level, the proportion of steps attempting multi-token generation increases from 47.1\% to 50.0\%, while the overall mean accepted length improves from 1.50 to 1.56 tokens. These findings suggest that Gaussian prediction parameters provide complementary information beyond the backbone context representation, improving both conditional acceptance and the effectiveness of adaptive generation.

\subsection{Generation Efficiency}
\label{sec:generation_quality_efficiency}

We evaluate the speedup relative to standard autoregressive decoding with Llama-3.1-8B-Instruct. Table~\ref{tab:alpaca_speedup} reports speedups of 1.28$\times$, 1.28$\times$, and 1.3$\times$ on Alpaca, GSM8K, and HumanEval, respectively. These results demonstrate that adaptive-granularity generation can preserve the autoregressive backbone's outputs while providing additional computational benefits. 

\begin{table}[h]
    \caption{Throughput speedup over greedy Llama}
    \label{tab:alpaca_speedup}
    \centering
    \small
    \setlength{\tabcolsep}{6pt}
    \adjustbox{max width=\columnwidth}{%
    \begin{tabular}{lccc}
        \toprule
        & \multicolumn{3}{c}{\textbf{Speedup}} \\
        \cmidrule(lr){2-4}
        \textbf{Model} & \textbf{Alpaca} & \textbf{GSM8K} & \textbf{HumanEval} \\
        \midrule
        Llama-3.1-8B-Instruct & $1.00\times$ & $1.00\times$ & $1.00\times$ \\
        CADENCE & $1.28\times$ & $1.28\times$ & $1.30\times$ \\
        \bottomrule
    \end{tabular}
    }
\end{table}

\section{Conclusion}

We introduced CADENCE, a framework for adaptive-granularity language generation that learns to vary the number of tokens predicted at each decoding step. By combining Gaussian future-span representations with a learned horizon-selection mechanism, CADENCE moves beyond the fixed one-token generation granularity of conventional autoregressive decoding. Experiments across instruction following, mathematical reasoning, and code generation demonstrate context-dependent variable-length generation, with longer accepted continuations frequently occurring in structured and predictable spans. Under greedy verification, CADENCE preserves the backbone's generated outputs while providing additional inference-efficiency benefits. Our findings suggest that learning when to generate multiple tokens is a promising direction toward more flexible language generation.

\section{Limitations}


First, while the learned router adapts prediction horizons to the context, its decisions can be conservative, occasionally selecting shorter horizons or backbone-only generation when longer continuations could be accepted.  Future work could explore more effective routing objectives that balance generation progress with computational cost. Second, our experiments are limited to a single backbone model, Llama-3.1-8B-Instruct. Finally, CADENCE relies on backbone verification to prevent errors in predicted continuations from propagating, and its behavior without verification is not systematically evaluated.

\bibliography{acl_references}

\appendix

\section{Additional Details}
\subsection{Exact-match agreement of router}
\label{sec:router-oracle}

We check whether the router selects the largest horizon that fully matches the backbone's greedy continuation.
On 80 Alpaca prompts, the router is scored against that label at every answer token (17{,}359 positions).
Horizon \(k\) is the backbone token plus \(k\) learned tokens, and label 0 means no horizon matches in full.
Table~\ref{tab:router-oracle} shows the pair at each position.

\begin{table}[h]
\centering
\small
\setlength{\tabcolsep}{4.5pt}
\renewcommand{\arraystretch}{1.15}
\begin{tabular}{lcccccc}
\toprule
& \multicolumn{6}{c}{Exact-match label} \\
\cmidrule(lr){2-7}
Router & 0 & 1 & 2 & 3 & 4 & 5 \\
\midrule
0 & \cellcolor{blue!50}\textbf{35.0} & \cellcolor{blue!25}12.3 & \cellcolor{blue!12}4.4 & \cellcolor{blue!4}1.1 & 0.3 & 0.1 \\
1 & \cellcolor{blue!11}3.9 & \cellcolor{blue!28}\textbf{13.8} & \cellcolor{blue!9}3.3 & \cellcolor{blue!4}1.0 & 0.3 & 0.1 \\
2 & \cellcolor{blue!7}2.4 & \cellcolor{blue!9}3.2 & \cellcolor{blue!16}\textbf{6.8} & \cellcolor{blue!5}1.5 & 0.4 & 0.1 \\
3 & \cellcolor{blue!4}1.1 & \cellcolor{blue!4}1.1 & \cellcolor{blue!4}1.0 & \cellcolor{blue!7}\textbf{2.1} & 0.4 & 0.1 \\
4 & 0.2 & 0.4 & 0.2 & 0.2 & \cellcolor{blue!4}\textbf{0.6} & 0.1 \\
5 & 0.4 & 0.3 & 0.3 & 0.4 & 0.3 & \cellcolor{blue!4}\textbf{0.8} \\
\bottomrule
\end{tabular}
\caption{Router prediction against the exact-match label for the UltraChat router, at every token of 80 Alpaca greedy answers (17{,}359 positions).
Each cell is the percentage of positions with that pair.
Class 0 is no full match.
Bold cells are exact agreement, and darker blue is a larger share of positions.}
\label{tab:router-oracle}
\end{table}

The router equals the label on 59.1\% of positions.
A predictor that always emits no-match would score 43.0\%, the label's own no-match rate, so the head is 16 points above that baseline.
The prediction is within one class of the label on 85.2\% of positions.
Of the disagreements, 25.4\% are shorter than the label and 15.5\% are longer.

Almost all of the mass is in three cells.
The router and the label agree on no-match (35.0\%) and on \(k{=}1\) (13.8\%).
The largest miss sits beside the first of these: the router says no-match while the label is \(k{=}1\) (12.3\%).
Horizons past \(k{=}2\) are rare both as predictions and as labels.

Agreement is lower on the first few answer tokens, where a longer full match is more common and the router still abstains.
Further into the answer the label is more often no-match, and agreement rises with it.

\label{sec:appendix}

\end{document}